\documentclass[11pt]{article}

\usepackage[final]{acl}
\usepackage{times}
\usepackage{latexsym}

\usepackage[T1]{fontenc}

\usepackage[utf8]{inputenc}
\usepackage{bm} 

\usepackage{microtype}

\usepackage{inconsolata}

\usepackage{graphicx}

\usepackage{hyperref}       
\usepackage{url}            
\usepackage{booktabs}       
\usepackage{amsfonts}       
\usepackage{nicefrac}       
\usepackage{microtype}      
\usepackage{xcolor}         
\usepackage{listings}
\usepackage{comment}
\usepackage{amsmath}
\usepackage{multirow}
\usepackage{adjustbox}
\usepackage[table]{xcolor}
\usepackage{threeparttable}
\usepackage{graphicx}
\usepackage{caption}
\usepackage{subcaption}
\usepackage{enumitem}
\usepackage[most]{tcolorbox}
\usepackage{lipsum} 
\usepackage{geometry}
\tcbuselibrary{breakable}
\usepackage{titlesec}
\usepackage{dblfloatfix}
\usepackage{soul}
\usepackage[ruled,vlined]{algorithm2e}
\usepackage{mathrsfs}

\definecolor{highlightcolor}{gray}{0.90}
\lstdefinestyle{sparql}{
  language=SQL, 
  basicstyle=\ttfamily\small,
  keywordstyle=\color{blue},
  stringstyle=\color{red},
  commentstyle=\color{gray},
  morekeywords={SELECT, WHERE}, 
  frame=single,
  breaklines=true,
  columns=fullflexible
}

\title{Topology-Aware Reasoning over Incomplete Knowledge Graph with Graph-Based Soft Prompting}

\author{
Shuai Wang$^{1}$ \quad
Xixi Wang$^{2}$ \quad
Yinan Yu$^{1}$ \\
$^{1}$Chalmers University of Technology and University of Gothenburg, Sweden \\
$^{2}$Technical University of Denmark, Kgs.\ Lyngby, Denmark \\
\texttt{\{shuaiwa,yinan\}@chalmers.se} \quad
\texttt{xixiwa@dtu.dk}
}

\begin{document}
\maketitle

\begin{abstract}
Though powerful, LLMs are still prone to hallucinations in knowledge-intensive tasks. Knowledge Base Question Answering (KBQA) mitigates this by grounding generation in Knowledge Graphs (KGs). But the inherently incomplete nature of knowledge graphs makes most multi-hop KBQA methods fragile due to their reliance on explicit edge traversal. In this paper, we proposed a novel graph-based soft prompting framework that shifts the reasoning paradigm from node-edge path traversal to subgraph-level reasoning. Specifically, we employ a Graph Neural Network (GNN) to encode extracted subgraphs into soft prompts, enabling LLM to reason over richer structural context and identify relevant entities beyond immediate graph neighbors, thereby reducing sensitivity to missing edges. Furthermore, we introduce a two-stage paradigm that reduces computational cost while preserving good performance: a lightweight LLM first leverages the soft prompts to identify question-relevant entities and relations, followed by a more powerful LLM for evidence-aware answer generation. Experiments on three incomplete benchmarks and four widely used KBQA benchmarks demonstrate the effectiveness of our framework. 
\footnote{Code is available at the repository: \url{https://github.com/Wangshuaiia/GraSP}.}
\end{abstract}

\section{Introduction}

Large Language Models (LLMs) have demonstrated remarkable capabilities across a wide range of applications. However, despite these advances, LLMs often suffer from hallucinations and unstable accuracy when handling knowledge-intensive tasks \cite{lavrinovics2025knowledge}. To address these limitations, Knowledge Graphs (KGs) have gained increasing attention as a reliable source of structured and explicit knowledge \cite{pan2024unifying}. By integrating the complementary strengths of LLMs and KGs, Knowledge Base Question Answering (KBQA) improves answer reliability through explicit retrieval and reasoning over entities and relations in knowledge graphs, enabling more accurate and evidence-grounded responses \cite{song2023advancements}. However, successful reasoning in KBQA often requires sufficiently connected reasoning paths within the KG.

\begin{figure}[t]
    \centering
    \includegraphics[width=\linewidth]{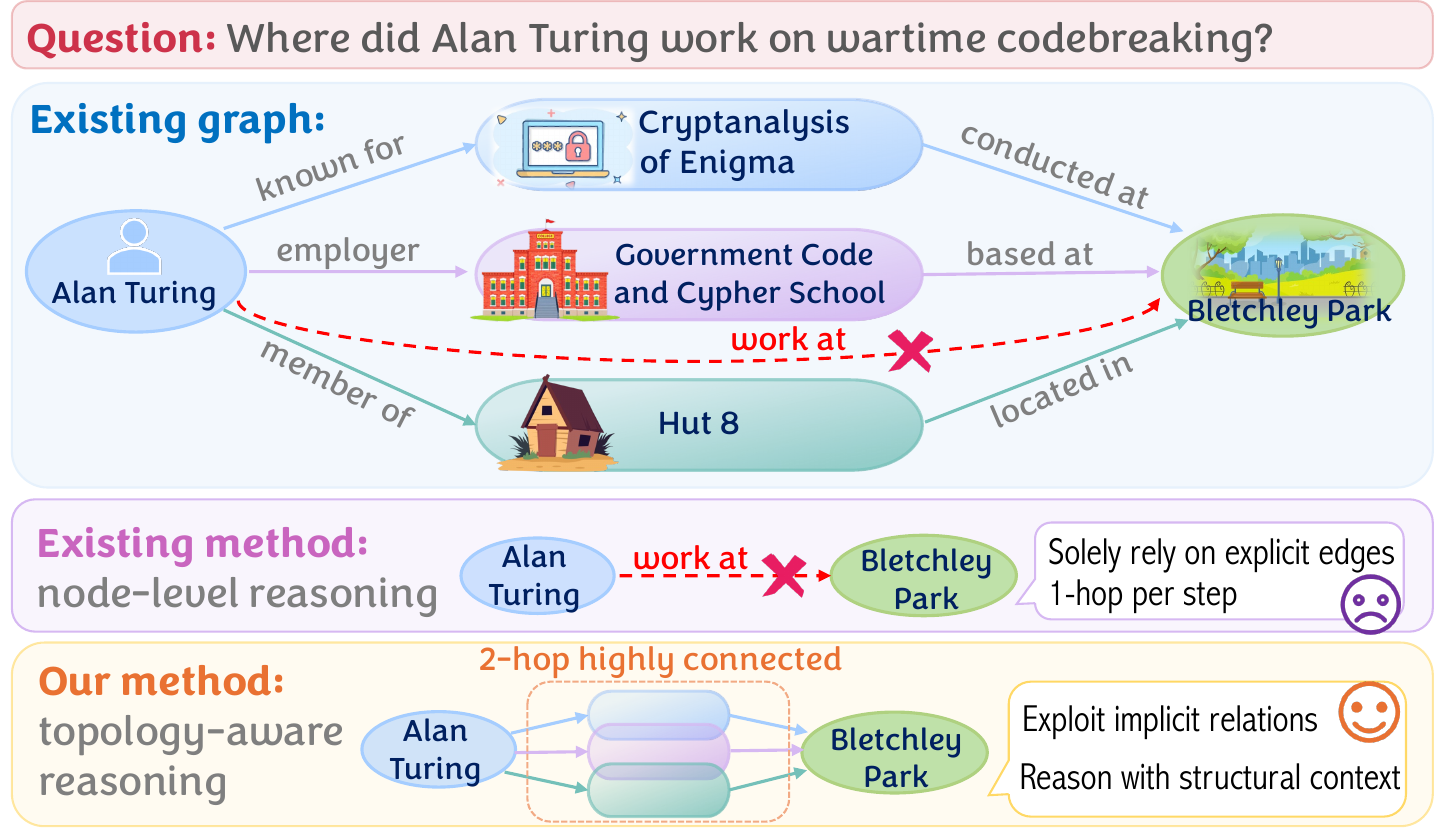}
    \caption{An example of reasoning over an incomplete knowledge graph. Existing methods (middle) solely rely on explicit edges and fail to perform multi-hop reasoning under graph incompleteness (e.g. missing edges). Our method (bottom) overcomes this limitation by exploiting implicit relations in the graph structure.}
    \label{fig:compare-intro}
\vspace{-0.5cm}
\end{figure}

Knowledge graphs are inherently incomplete due to limitations in both their construction and maintenance processes \cite{liu2026debate,xu2024generate}. During construction, information extraction errors, entity ambiguity, ontology limitations, and long-tail knowledge sparsity prevent many real-world facts from being incorporated into the graph. During maintenance, the continuously evolving nature of real-world knowledge further introduces missing relations. However, existing KBQA methods are often strictly confined to explicit KG edges \cite{ma2025large}, making the reasoning process highly sensitive to missing relations that can easily disrupt multi-hop inference. Consequently, KG incompleteness poses significant challenges to KBQA, particularly for multi-hop ones.


As illustrated in the Figure \ref{fig:compare-intro}, answering ``\textit{Where did Alan Turing work on wartime codebreaking?}'' requires retrieving entities related to Alan Turing’s workplace. The absence of a direct edge between \textit{Alan Turing} and \textit{Bletchley Park} blocks further forward expansion, causing the reasoning chain to break.

Our key insight is that even in the absence of explicit edges, the surrounding subgraph structure often provides sufficient evidence to bridge reasoning gaps. \citet{wan2025digest} attempt to enrich contextual information by extracting subgraphs and converting them into textual descriptions via LLMs. Although this graph-to-text transformation provides additional contextual evidence for missing relations, it inevitably obscures the original graph topology, weakening the model’s ability to capture implicit structural dependencies between seemingly disconnected entities. However, the topological structure of KGs often contains essential information \cite{sun2018rotate}. 
For example, as illustrated in the Figure \ref{fig:compare-intro}, although \textit{Alan Turing} and \textit{Bletchley Park} are not directly connected, their local subgraph exposes convergent associative paths, such as (Alan Turing, member of, Hut 8) $\rightarrow$ (Hut 8, located in, Bletchley Park). Such structural patterns offer strong structural cues for inferring the missing relation \textit{worked at} between \textit{Alan Turing} and \textit{Bletchley Park}.

Graph Neural Networks (GNNs) are well suited to exploit this information by capturing complex entity dependencies through message-passing mechanisms \cite{wu2020comprehensive}. Unlike previous methods that make isolated, step-by-step decisions, using GNNs allows candidate entities to be evaluated within their local structural contexts, enabling the LLM to reason beyond immediate neighbors and reach entities without direct connections.




In this paper, we proposed a graph-based soft prompting (\textbf{GraSP}) framework for multi-hop KBQA. GraSP first employs a GNN to encode the retrieved subgraph, capturing its rich structural dependencies as dense entity representations. We then used a transformation layer to map these representations into soft prompts, which provides the LLM with global structural context. Inspired by the two-stage paradigm of \citet{chen2023accelerating}, we first used a small LLM to identifies question-relevant entities in the subgraph with the help of the soft prompts. Then we further injected the selected entities together with their associated relations into another larger LLM to generate accurate and high-quality answers. This two-stage approach, using a lightweight LLM to generate a draft and then a larger LLM to refine and verify, improves overall performance while reducing computational cost.
Our contributions are summarized as follows: 

\begin{itemize}[leftmargin=*]
\item We propose a novel subgraph-level reasoning paradigm that encodes structural subgraphs into graph-based soft prompts using a GNN, enabling more robust reasoning under knowledge graph incompleteness.
\item We propose a two-stage reasoning paradigm in which a lightweight model performs soft-prompt-based reasoning, and a powerful LLM generates evidence-aware answers, reducing computational cost while enabling seamless integration with off-the-shelf LLMs.
\item Extensive experiments on three incomplete KG benchmarks and four widely used KBQA benchmarks demonstrate the effectiveness of our framework.
\end{itemize}

\section{Related Work}
\label{sec:related-work}
\subsection{Knowledge Graph Reasoning}
Multi-hop knowledge base question answering (KGQA) requires reasoning over multiple relations to find answers. 
Recent work increasingly treats LLMs as reasoning agents that explore knowledge graphs. Methods such as Think-on-Graph \cite{sun2024thinkongraph} perform structured traversal over entities and relations, while Plan-on-Graph \cite{chen2024plan} and Debate-on-Graph \cite{ma2025debate} introduce step-wise verification or role-based collaboration to reduce spurious reasoning. Other approaches enhance exploration via auxiliary mechanisms, including memory augmentation \cite{jiang-etal-2025-kg} and iterative query decomposition \cite{wang-yu-2025-iquest,huang2023question}. 
Learning-based methods further guide multi-hop reasoning, including transition modeling \cite{zhao2024kg} and reinforcement learning–based decision processes \cite{zhang2025collaborative}. To improve efficiency, LightPROF \cite{ao2025lightprof} compresses reasoning graphs into compact embeddings. However, these approaches still rely on explicit KG edges at inference time. LMP \cite{wan2025digest} alleviates this by linearizing subgraphs into text, but with a loss of graph structure.

\subsection{Incomplete Knowledge Graphs}

Incomplete knowledge graphs are traditionally addressed by knowledge graph completion (KGC), which employs embedding-based, structure-aware, or large pretrained models to predict missing links \cite{chao-etal-2021-pairre,zhu2021nbfnet,saxena-etal-2022-kgt5}. Some works further explores inductive and foundation-model-based KGC to improve transferability \cite{galkin2023ultra,saxena-etal-2022-kgt5,hua2025merry}. Despite methodological advances, these methods are primarily designed to explicitly complete the KG before downstream use. 

There are also some work focuses on reasoning over incomplete KGs. To mitigate missing knowledge at inference time, \citet{guo2023knowledge} estimate the likelihood of missing relations in the retrieved subgraph, but this approach still relies on explicit edge prediction and makes limited use of the global graph structure. 
Alternatively, Generate-on-Graph \cite{xu2024generate} employs LLMs to generate implicit facts, but such generation may suffer from hallucination. 
\citet{liu2026debate} further supplement missing KG information by retrieving external documents from the web.
In contrast, our work perform multi-hop KGQA directly over incomplete graphs, without requiring prior graph completion.

\subsection{Soft Prompt Reasoning}

Soft prompting enables parameter-efficient adaptation of frozen language models by optimizing continuous prompt vectors \cite{li2021prefix,lester2021power,liu2022ptuningv2}. Recently, this paradigm has been extended to latent reasoning \cite{xu2025softcot,hao2024coconut,zhang2025softthinking} as well as structured reasoning, like graph-based reasoning \cite{liu2024graphprompter,tian2024graphneuralprompt,yang2024agot}. However, existing methods typically require access to model internals, which restricts their applicability to open-source LLMs. In contrast, our proposed two-stage paradigm applies soft prompting to a lightweight open-source model and then translates the learned representations into natural language, enabling seamless integration with any off-the-shelf closed-source LLMs.

\section{Task Definition}
We consider the problem of Knowledge Base Question Answering (KBQA) over a knowledge graph 
\( G \subseteq \mathcal{E} \times \mathcal{R} \times \mathcal{E} \), 
where \( \mathcal{E} \) and \( \mathcal{R} \) denote the sets of entities and relations, respectively. 
Each fact in the graph is represented as a triple \( t = (e, r, e') \), indicating that a relation \( r \) holds between entities \( e \) and \( e' \). 
Given a natural language question \( q \) and a topic entity \( e_t \in \mathcal{E} \) identified from the question, 
the goal is to determine the corresponding answer entity \( e_a \in \mathcal{E} \) by reasoning over \( G \).  This process typically involves traversing the graph starting from \( e_t \). 
However, due to the sparsity and incompleteness of real-world knowledge graphs, 
the answer entity often cannot be reached via direct one-hop relations. 
Consequently, effective KBQA systems must support multi-hop reasoning, 
where \( e_a \) is connected to \( e_t \) through a sequence of intermediate entities and relations.

\begin{figure*}[t]
    \centering
    \includegraphics[width=\linewidth]{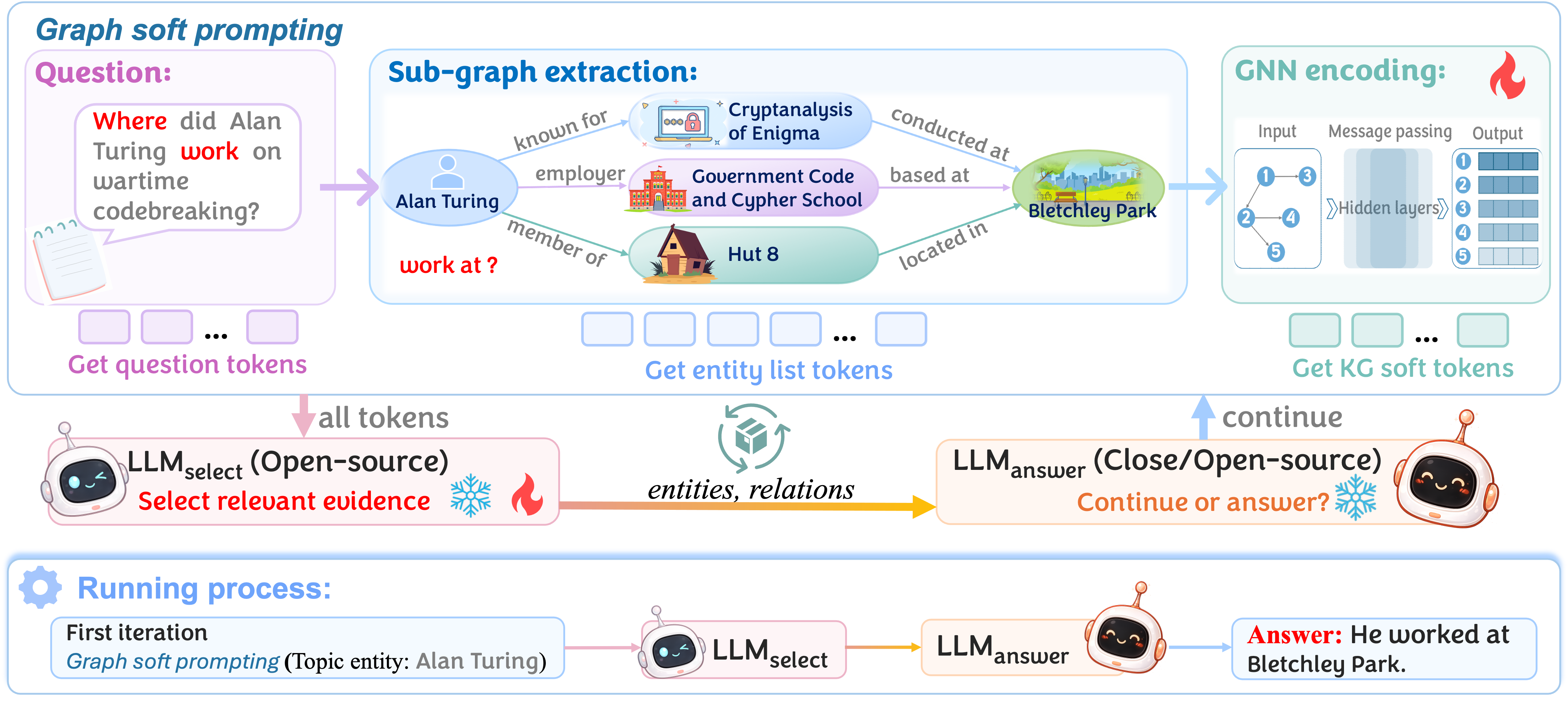}
    \caption{Overview of the proposed framework and its running process. The upper part describes the method: GraSP is used to construct structure-aware soft prompts from the sub knowledge graph, which are first processed by a compact LLM to select relevant entities and relations, and then refined by a more powerful LLM with substantially more parameters to generate high-quality answers. The lower part illustrates an example of the reasoning process that obtains the answer in a single iteration.}
    \label{fig:compare-method}
    \vspace{-0.2cm}
\end{figure*}

\section{Method}
Our proposed method is illustrated in Figure~\ref{fig:compare-method}. It consists of three components: (1) \textit{Subgraph Extraction}, (2) \textit{Graph-based Soft Prompting}, and (3) \textit{Two-Stage Reasoning and Answering}. Given a complex query, we first extract a subgraph centered on the topic entities and encode its structure using a GNN with message passing. The encoded entity representations are then used to construct soft prompts. Finally, we decouple reasoning and answering via a two-stage LLM framework, reducing computational cost while enabling flexible integration with off-the-shelf LLMs.

\subsection{Subgraph Extraction}
\label{sec:subgraph-extract}

We begin subgraph extraction based on a specific entity, which is the topic entity when reasoning starts. 
We aim to retrieve its \( l \)-hop neighborhood of the entity; however, directly expanding all neighbors is infeasible due to the large number of irrelevant entities. To address this, we adopt a relevance-based retrieval strategy.

Specifically, we encode the question \( q \) and each candidate triple \(t\) with BERT that use the embedding of [cls] token as representation, and then compute their cosine similarity:
\begin{equation}
s(q, t) = \cos(\mathbf{q}, \mathbf{t}),
\end{equation}
where \( \mathbf{q} \) and \( \mathbf{e} \) denote the vector representations of the question and the entity, respectively. Entities are ranked according to \( s(q, t) \), and the top-\( k_1 \) entities are selected.

To enable multi-hop expansion, we iteratively repeat this process: for each selected entity, we retrieve its neighboring entities and again select the top-\( k_2 \) entities based on similarity to the question. By iterating this procedure up to \( l \) hops, we obtain a question-relevant \( l \)-hop subgraph. The hop number \( l \) is treated as a hyperparameter. The specific SPARQL queries used for neighbor retrieval are provided in the Appendix \ref{sec:sparql}.

\subsection{Graph-based Soft Prompting}
\label{sec-grasp}
To encode the extracted subgraph, we employ an \( l \)-layer graph neural network. Since cosine similarity alone is insufficient to model relational dependencies, we adopt a graph attention network to focus on entities that are more relevant to the question.

At each layer, for a target entity \( e_i \) at $(l+1)$-th layer and its neighbor \( e_j \) at $l$-th layer, let \( \mathbf{h}^{(l+1)}_i \in \mathbb{R}^{d}\) and \( \mathbf{h}^{(l)}_j\in \mathbb{R}^{d}\) denote their representations, respectively. The node update is computed as:
\begin{equation}
\mathbf{h}^{(l+1)}_i =
\sigma \Bigl(
\sum_{j} a_{ij} \mathbf{W}^{(l)} \mathbf{h}^{(l)}_j + \mathbf{b}^{(l)}
\Bigr),
\end{equation}
where \( \mathbf{W}^{(l)} \in \mathbb{R}^{d\times d}\) and \( \mathbf{b}^{(l)}\in \mathbb{R}^{d} \) are learnable parameters, $\sigma$ represents an activation function, and \( a_{ij} \) is the attention weight between entities \( e_i \) and \( e_j \).

The attention weight \( a_{ij} \) is computed as:
\begin{equation}
\resizebox{0.89\linewidth}{!}{$
a_{ij} =
\text{softmax}\left(
f\big(
\mathbf{W}(\mathbf{q}\,\|\,\mathbf{h}_i),
\mathbf{W}(\mathbf{h}_j\,\|\,\mathbf{r}_{ij})
\big)
\right),
$}
\end{equation}
where \( f\) is the scoring function using inner production (i.e., \(\mathbb{R}^d \times \mathbb{R}^d \rightarrow \mathbb{R} \)), \( \mathbf{r}_{ij} \) denotes the relation embedding between \( e_i \) and \( e_j \), and \( \| \) represents vector concatenation.

The GNN layer computes the message passing over the entire graph structure as:
\begin{equation}
\resizebox{0.89\linewidth}{!}{$
\mathbf{H}^{(l+1)} =
\sigma \Bigl(
\mathbf{D}^{-1/2}\hat{\mathbf{A}}\mathbf{D}^{-1/2}
\mathbf{H}^{(l)}\mathbf{W}^{(l)}
+ \mathbf{b}^{(l)}
\Bigr),
$}
\end{equation}
where $\mathbf{H}\in \mathbb{R}^{n\times d}$ denotes the representation matrix of $n$ entities, 
$\mathbf{D}\in \mathbb{R}^{n\times n}$ is the degree matrix which its diagonal element represent how many entities connected to each entity, and 
$\hat{\mathbf{A}} = [a_{ij}]$  denotes the adjacency matrix.
After \( l \) layers, we obtain contextualized embeddings for all entities in the subgraph.

We then transform the embeddings of all entities in the subgraph into a soft prompt. 
To bridge the modality gap between graph structures and natural language, we introduce a lightweight feed-forward network (FFN) to project these representations into the embedding space of the LLM:

\begin{equation}
\mathbf{\hat{H}} =
\text{FFN}(\mathbf{H}),
\end{equation}
where \( \mathbf{\hat{H}} \in \mathbb{R}^{n\times d_{\text{LLM}}}\) represents the aggregated graph-level soft prompt matrix.

We then concatenate the LLM embeddings of instruction context ${I}_{\text{instr}}$, the question $q$, and the list of candidate entities ${E}$ with the soft prompt \( \mathbf{\hat{H}} \), and then feed them input to the LLM.
\begin{equation}
\mathbf{x}_{\text{LLM}} =
\text{concat}[\textbf{I}_{\text{instr}}, \textbf{q}, \textbf{E}, \mathbf{\hat{H}}],
\end{equation}
All textual information together with the graph-based soft prompt constitutes the final prompt, guiding the LLM to select entities relevant to the question.

\subsection{Two-Stage Reasoning and Answering}
To reduce computational cost while maintaining strong performance, we borrowed the conception of draft-and-refine framework~\citet{chen2023accelerating} and utilize a two-stage answering framework.

First, the generated soft prompt is fed into a lightweight LLM with fewer parameters, which is used to extract the entity (or entities) most relevant for answering the question:
\begin{equation}
{\hat{E}}=\text{LLM}_{\text{select}}(\mathbf{x}_{\text{LLM}})
\end{equation}

To efficiently leverage powerful closed-source models, we design the output of the first stage in natural language, allowing it to serve as explicit evidence that can be jointly consumed with the question by any powerful off-the-shelf LLM for answer generation. 
Instead of directly generating an answer, the LLM first assesses whether the current evidence is sufficient. If not, it selects an intermediate entity as the next reasoning target and repeats the subgraph extraction and reasoning process. Once the predefined iteration limit is reached, the model generates the answer based on the available evidence and its internal knowledge.
Formally, this decision process is defined as:
\begin{equation}
\resizebox{0.89\linewidth}{!}{$
\text{LLM}_{\text{ans}}(q, \hat{E}, \hat{R}) =
\begin{cases}
\text{answer}, &
\shortstack{\text{If sufficient}\\\text{or limit reached}}, \\
\text{entity}, & \text{otherwise}.
\end{cases}
$}
\end{equation}
Here, ${\hat{R}}$ denotes the set of relations corresponding to the entity set ${\hat{E}}$.

\subsection{Optimization}

Overall, the trainable components of our GraSP framework include the GNN and the FFN in Section~\ref{sec-grasp}.

For the LLM used in entity selection (denoted as \( \text{LLM}_{\text{select}} \) in Section~5.3), we allow two training strategies: either fine-tuning the model or keeping it frozen. Since the primary objective of this component is to identify question-relevant entities, we optimize it using a cross-entropy loss:
\begin{equation}
\mathcal{L} = - \log P(y \mid \mathbf{I}_{\text{instr}}, \mathbf{Q}, \mathbf{E}, \mathbf{\hat{H}}),
\label{func:loss}
\end{equation}
where \( y \) denotes the ground-truth target entity. Gradients are backpropagated through the graph-based soft prompt \( \mathbf{\hat{H}} \) to update the parameters of both the GNN and the FFN.

In contrast, for $\text{LLM}_{\text{ans}} $ in the answer generation stage, once sufficient information has been retrieved, it answers the question does not rely on task-specific modeling assumptions. Therefore, the LLM used for answer generation can be any off-the-shelf LLM and does not require fine-tuning.

\begin{table*}[t]
\centering
\small
\caption{Performance comparison on three incomplete KGQA datasets (Hits@1). We re-implemented all baselines using GPT-5.2 for fair comparison. ``$_{\pm}$'' denotes the standard variation. ``$^{*}$'' ($p<0.05$), ``$^{**}$'' ($p<0.01$), and ``$^{***}$'' ($p<0.001$) denote statistically significant improvements over LMP under the \textit{t}-test.
}
\renewcommand{\arraystretch}{1.1}
\label{tab:ikgqa_results}
\begin{tabular}{lccc}
\toprule
Method & IKGWQ & CWQ-IKG & WebQSP-IKG \\
\midrule
ToG~\cite{sun2024thinkongraph} & 58.6$_{\pm2.4}$ & 59.2$_{\pm2.3}$ & 70.2$_{\pm1.7}$ \\
DoM \cite{liu2026debate} & 59.1$_{\pm1.3}$ & 60.7$_{\pm1.5}$ & 75.8$_{\pm1.8}$ \\
GoG \cite{xu2024generate} & 60.3$_{\pm3.4}$ & 61.4$_{\pm2.7}$ & 78.1$_{\pm2.4}$ \\
iQUEST~\cite{wang-yu-2025-iquest} & 60.1$_{\pm1.6}$ & 59.7$_{\pm1.4}$ & 79.3$_{\pm1.2}$ \\
KBQA-o1~\cite{luokbqa} & 61.1$_{\pm1.2}$ & 63.9$_{\pm2.6}$ & 80.1$_{\pm0.9}$ \\
LMP~\cite{wan2025digest} & 62.7$_{\pm2.2}$ & 64.3$_{\pm1.4}$ & 80.6$_{\pm1.7}$ \\
\midrule
\textbf{GraSP (Ours)} & \textbf{65.5}$_{\pm1.4}$$^{*}$ & \textbf{68.6}$_{\pm1.0}$$^{***}$ & \textbf{83.9}$_{\pm1.1}$$^{**}$ \\
\bottomrule
\end{tabular}
\end{table*}

\section{Experiments}
\subsection{Datasets}
\label{sec:datasets-experimens}
We conducted experiments on three incomplete KG datasets and four widely used KBQA datasets. The first incomplete KG dataset is IKGWQ \cite{liu2026debate}, which is constructed based on real-world knowledge updates missing from outdated knowledge graphs. The questions and answers are rebuilt using newly emerged facts, making the benchmark more realistic for evaluating hybrid reasoning over both KGs and external knowledge sources.
We also used the incomplete KG settings proposed by \citet{xu2024generate}, which simulate incompleteness by randomly removing triples (40\%) from CWQ and WebQSP. We denoted the datasets as CWQ-IKG and WebQSP-IKG, repsectively.

In addition, we evaluated our approach on four widely-used datasets for KBQA, leveraging a large-scale general-purpose knowledge graphs Freebase: ComplexWebQuestions (CWQ)~\cite{talmor2018web}, WebQuestionsSP (WebQSP)~\cite{yih2016value}, WebQuestions~\cite{berant2013semantic}, and GrailQA~\cite{gu2021beyond}. 

The characteristics of all datasets are provided in Table \ref{tab:datasets} in Appendix \ref{app-datasets}.
Following prior work~\cite{sun2024thinkongraph, zhao2024kg, xiong-etal-2024-interactive}, we used Hit@1 score as the evaluation metrics. The implementation details are provided in Appendix \ref{sec:impl}.

\subsection{Comparative Methods}

We compared our approach with the following state-of-the-art methods:

\begin{itemize}[leftmargin=*, itemsep=2pt]
    \item \textbf{ToG}~\cite{sun2024thinkongraph}: A step-by-step reasoning framework based on LLMs. 

\item \textbf{DoM} \cite{liu2026debate}: Uses multiple agents to integrate structured and unstructured knowledge for incomplete KG reasoning. We exclude external documents for fair comparison.

\item \textbf{GoG} \cite{xu2024generate}: Uses LLMs to generate question-related edges or entities to support reasoning over incomplete KGs.
    
    
    \item \textbf{iQUEST}~\cite{wang-yu-2025-iquest}: Performs knowledge graph reasoning by iteratively generating and answering sub-questions.

    \item \textbf{KBQA-o1}~\cite{luokbqa}: Enhances search efficiency on knowledge graphs by training a Monte Carlo Tree Search strategy, which requires model fine-tuning. We re-implement this based on LLaMA3.3-70B.
        
    \item \textbf{LMP}~\cite{wan2025digest}: Extracts entity-centered first-order subgraphs and converts them into natural language explanations using an LLM, providing textual evidence for reasoning.
\end{itemize}

\subsection{Main Results}
\label{sec:main-results}


\paragraph{Results on Three Incomplete KG Datasets}
Table~\ref{tab:ikgqa_results} presents the results on three incomplete KG benchmarks. For a fair comparison, all baseline methods were unified under the same backbone model, GPT-5.2. Our method achieves the best performance across all three datasets. To ensure reliability, each experiment was repeated five times and the standard deviations are reported.

To further verify the stability of the improvements, we conducted paired t-tests against the strongest baseline, LMP. The results show that our method significantly outperforms LMP on IKGWQ ($p<0.05$), CWQ ($p<0.001$), and WebQSP ($p<0.01$). These results demonstrate the consistent superiority of our framework over existing baselines. We attribute the improvements to our ability to better exploit the topological structure of knowledge graphs, which helps alleviate the impact of KG incompleteness.

\paragraph{Results on Four Widely Used KBQA Datasets}
We provide the results on four widely used KBQA datasets in Table~\ref{tab:kbqa_results_four_widely_used} in Appendix \ref{sec:additional_results_four_widely_used}. Our approach achieves the best performance on three multi-hop datasets, even when compared with very strong baselines. These results demonstrate that our method effectively leverages graph structure through a soft-prompt-based design and parallel GNN-based graph representation learning. Such a design is particularly beneficial for multi-hop reasoning under KG incompleteness, where missing information may disrupt reasoning paths.

\begin{table*}[t]
\centering
\small
\caption{Ablation study of the graph-based soft prompting.}
\renewcommand{\arraystretch}{1.1}
\resizebox{\textwidth}{!}{
\begin{tabular}{lccccccc}
\toprule
\multirow{2}{*}[-0.5ex]{\textbf{Method}}
& \multicolumn{3}{c}{{Incomplete KG Benchmarks}} 
& \multicolumn{4}{c}{{Standard KBQA Benchmarks}} \\
\cmidrule(lr){2-4} \cmidrule(lr){5-8}
& \textbf{IKGWQ} & \textbf{CWQ-IKG} & \textbf{WebQSP-IKG} 
& \textbf{CWQ} & \textbf{WebQSP} & \textbf{WebQuestion} & \textbf{GrailQA} \\
\midrule
\textbf{GraSP (GPT-20B as $\text{LLM}_{\text{select}}$)} & 65.52 & 68.57 & 83.91 & 82.58 & 93.63 & 84.32 & 83.99 \\
\quad Without Soft Prompt & 61.68 & 63.92 & 79.64 & 78.72 & 89.42 & 81.28 & 78.46 \\
\quad Triplet-based Representation & 62.37 & 65.41 & 81.26 & 79.63 & 92.13 & 82.26 & 80.74 \\
\quad Text-based Soft Prompt & 63.18 & 66.02 & 82.17 & 80.15 & 92.41 & 83.55 & 81.63 \\
\midrule
\textbf{GraSP (Qwen-30B-A3B as $\text{LLM}_{\text{select}}$)} & 64.42 & 67.12 & 81.23 & 82.28 & 93.16 & 84.07 & 83.46 \\
\quad Without Soft Prompt & 60.37 & 62.48 & 76.95 & 77.94 & 88.05 & 82.29 & 79.21 \\
\quad Triplet-based Representation & 62.06 & 64.71 & 78.83 & 80.23 & 91.17 & 82.10 & 80.16 \\
\quad Text-based Soft Prompt & 62.78 & 65.36 & 79.64 & 80.53 & 91.41 & 82.18 & 81.27 \\
\bottomrule
\end{tabular}
}
\label{tab:gnn_ablation}
\vspace{-0.4cm}
\end{table*}

\subsection{Ablation on Graph-based Soft Prompt}

To investigate in detail the effect of the Graph-based Soft prompt, we designed several variants for comparison. (1) \textit{Without Soft Prompt}, we removed the soft prompting module and allowed the LLM to directly select question-relevant entities. (2) \textit{Triplet-based Representation}, where the retrieved subgraph was linearized into a list of knowledge triplets, which were then provided to the LLM as input, discarding the explicit graph structure. (3) \textit{Text-based Soft Prompt}, in which all triplets were encoded into semantic vectors using BERT, and the resulting embeddings were used as soft prompts instead of graph-structured representations. 

The results are shown in Table~\ref{tab:gnn_ablation}. Overall, GraSP achieves the best performance across all settings. Removing graph information (\textit{Without Soft Prompt}) causes a significant performance drop, showing the importance of structured knowledge for multi-hop reasoning. \textit{Triplet-based Representation} improves over the no-graph baseline, while \textit{Text-based Soft Prompt} further enhances performance, demonstrating the effectiveness of soft prompts. Finally, \textit{Graph-based Soft Prompt} consistently outperforms \textit{Text-based Soft Prompt}, indicating that GNNs can better capture structural information. The trends are consistent across different LLMs, verifying the effectiveness and robustness of GraSP.

\subsection{Performance under Varying Degrees of KG Incompleteness}
\begin{figure}[t]
    \centering
    \includegraphics[width=1\linewidth]{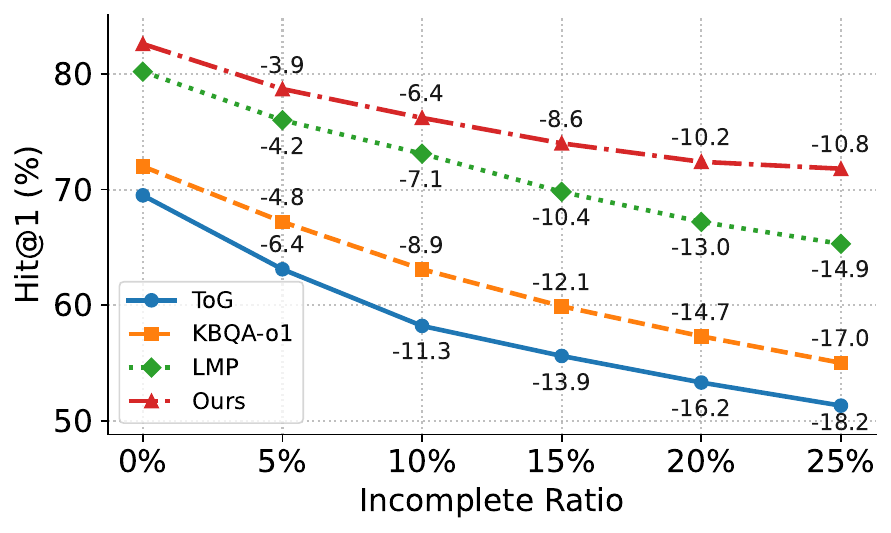}
    \caption{Results on CWQ under Knowledge Graph Incompleteness.}
    \label{fig:backtrack}
\vspace{-0.5cm}
\end{figure}

To evaluate the robustness of our method under varying degrees of KG incompleteness, we follow \citet{xu2024generate} and conduct experiments on the challenging CWQ dataset by randomly removing a portion of edges connected to topic entities. This setting simulates different levels of KG incompleteness. The removal ratio ranges from 5\% to 25\% to assess the impact of increasing incompleteness. We compared our approach with three representative baselines, ToG, KBQA-o1 and LMP. 
Notably, LMP addresses KG incompleteness by converting subgraphs into redundant natural language representations, which partially alleviates the impact of missing edges.
The results are shown in Figure~\ref{fig:backtrack}.

As the proportion of removed edges increases, all methods exhibit performance degradation compared to using the complete KG, with more severe drops under higher levels of incompleteness, which is expected. However, across all incompleteness settings, our method consistently shows a slower degradation trend than the baselines. In particular, the overall performance drop of our approach is only 10.8 points, significantly smaller than that of ToG (18.2), KBQA-o1 (17.2) and LMP (14.9). 
While both GraSP and LMP are more effective at mitigating the impact of KG incompleteness, ToG and KBQA-o1 suffer substantially larger performance degradation, as they rely heavily on explicit KG edge traversal for reasoning. In contrast, our method performs reasoning over a retrieved subgraph augmented with graph-based soft prompts, which effectively mitigates the impact of missing edges. These results demonstrate that GraSP is more robust to KG incompleteness.

\begin{table*}[t]
\centering
\small
\vspace{-0.4cm}
\caption{Performance comparison with different $\text{LLM}_{\text{select}}$ settings.}
\renewcommand{\arraystretch}{1.1}
\resizebox{\textwidth}{!}{
\begin{tabular}{lccccccc}
\toprule
\multirow{2}{*}[-0.5ex]{\textbf{Method}}
& \multicolumn{3}{c}{{Incomplete KG Benchmarks}} 
& \multicolumn{4}{c}{{Standard KBQA Benchmarks}} \\
\cmidrule(lr){2-4} \cmidrule(lr){5-8}
& \textbf{IKGWQ} & \textbf{CWQ-IKG} & \textbf{WebQSP-IKG} 
& \textbf{CWQ} & \textbf{WebQSP} & \textbf{WebQuestion} & \textbf{GrailQA} \\
\midrule

Qwen3-0.6B (freeze) & 48.36 & 51.42 & 66.75 & 61.73 & 69.14 & 58.05 & 62.48 \\

Qwen3-0.6B (Finetune) & 58.21 (+9.85) & 60.93 (+9.51) & 77.46 (+10.71) & 71.31 (+9.58) & 81.14 (+12.00) & 68.52 (+10.47) & 69.87 (+7.39) \\

\midrule

Qwen3-8B (freeze) & 57.84 & 60.91 & 77.62 & 72.55 & 81.54 & 72.31 & 65.38 \\

Qwen3-8B (Finetune) & 64.13 (+6.29) & 67.39 (+6.48) & 82.48 (+4.86) & 80.85 (+8.30) & 88.86 (+7.32) & 78.40 (+6.09) & 79.16 (+13.78) \\

\midrule

GPT-20B (freeze) & 60.08 & 63.24 & 78.37 & 77.21 & 86.34 & 78.28 & 76.13 \\

GPT-20B (fine-tune) & 65.52 (+5.44) & 68.57 (+5.33) & 83.91 (+5.54) & 82.58 (+5.37) & 93.63 (+7.29) & 84.32 (+6.04) & 83.99 (+7.86) \\

\bottomrule
\end{tabular}
}
\label{tab:finetune_gain}
\end{table*}

\begin{table}[t]
\centering
\small
\caption{Performance comparison of different hop settings on four standard benchmarks.}
\renewcommand{\arraystretch}{1.1}
\begin{tabular}{lcccc}
\toprule
& CWQ & WebQSP & WebQuestion & GrailQA \\
\midrule
1-Hop & 79.07 & 90.87 & 82.01 & 81.29 \\
2-Hop & 82.58 & 93.63 & 84.32 & 83.99 \\
3-Hop  & 76.47 & 85.98 & 80.21 & 78.66 \\
\bottomrule
\end{tabular}
\vspace{-0.4cm}
\label{tab:hop_ablation}
\end{table}

\subsection{Effectiveness of Two-Stage Reasoning}
Firstly, we evaluate the impact of the compact model $\text{LLM}_{\text{select}}$ in the first stage. Specifically, we experimented with LLMs of different scales, including Qwen3-0.6B, Qwen3-8B and GPT-20B, and further examined the effect of fine-tuning. We fix $\text{LLM}_{\text{answer}}$ as GPT-5.2 across all experiments. The results are summarized in Table~\ref{tab:finetune_gain}.

Overall, larger $\text{LLM}{\text{select}}$ models achieve better performance after fine-tuning. Notably, once fine-tuned, even an 8B model can achieve strong performance comparable to GPT-20B, suggesting that a tuned $\text{LLM}{\text{select}}$ collaborates more effectively with the GNN to identify relevant entities. Meanwhile, we observe that, benefiting from our paradigm, the performance remains acceptable even without fine-tuning; under this setting, the 0.6B model still delivers competitive results.

We further extended our evaluation to a broader range of LLMs, with results reported in Table~\ref{tab:different_select_llm} (Appendix~\ref{app-two-stage}).The results consistently show that our method remains stable even when using very small size LLMs. 
This is because LLMs are more effective at verification and refinement, making the framework more tolerant to noise introduced in the selection stage. 

To examine the impact of the answer LLM used in the second stage, we experimented with seven different LLMs. The results are reported in Table~\ref{tab:answer_llm_results} (Appendix~\ref{app-two-stage}). As shown, performance consistently improves as the model scale increases, which aligns well with our expectations. This trend further demonstrates that our two-stage framework can effectively leverage powerful off-the-shelf LLMs, highlighting a key practical advantage of the proposed approach.



\subsection{Impact of Subgraph Structure}
As described in Section \ref{sec:subgraph-extract}, subgraph extraction starts from a topic entity and collects its $l$-hop neighbors to construct an induced subgraph. Increasing the hop number introduces more multi-hop neighbors, resulting in a structurally more complex subgraph. To examine the impact of the hop number on subgraph construction, we evaluate our method using 1-hop, 2-hop, and 3-hop subgraphs, with the results reported in Table~\ref{tab:hop_ablation}.

The results show that 2-hop subgraphs achieve the best overall performance. With 1-hop extraction, missing edges cannot be effectively compensated, making it difficult to infer potential connections between entities; besides, the resulting subgraph is often too simple to fully exploit the proposed method. In contrast, 2-hop subgraphs can partially recover missing relations through intermediate entities, as the example in Figure \ref{fig:compare-intro}. However, increasing the hop number to 3 introduces excessive entities and edges, which complicate information passing and introduces substantial noise, leading to little benefit. Therefore, we adopt 2-hop subgraphs as they strike a favorable balance between structural expressiveness and noise, achieving effective encoding of graph-based soft prompts.



\subsection{Computational Efficiency Analysis}
To evaluate the computational efficiency of our approach, we compared its runtime and LLM invocation cost with those of ToG and LMP across different hop settings. For a fair comparison and to eliminate the influence of failure cases, we restricted the analysis to instances where all methods successfully complete the reasoning process. The results are summarized in Table~\ref{tab:runtime_hop}. As shown, our method consistently achieves the lowest runtime across all settings. Moreover, owing to the proposed two-stage reasoning paradigm, the number of calls to large-scale LLMs is substantially reduced, which significantly improves overall inference efficiency.

\begin{table}[t]
\centering
\caption{Comparison of LLM call counts and runtime across different hop settings for our method and baseline approaches. For our method, calls to the lightweight LLM are shown in \textcolor{green}{green}, while calls to the powerful LLM are shown in \textcolor{red}{red}.}
\label{tab:runtime_hop}
\small
\begin{tabular}{l l c c c}
\toprule
Method & Metrics & 1-Hop & 2-Hop & 3-Hop \\
\midrule
\multirow{2}{*}{ToG} 
 & \# LLM call  & 2.2   & 5.6   & 8.3  \\
 & Run time (s) & 3.16  & 11.76 & 17.51 \\
\midrule
\multirow{2}{*}{LMP} 
 & \# LLM call  & 2.3   & 4.9   & 7.3  \\
 & Run time (s) & 3.45  & 9.76  & 15.38 \\
\midrule
\multirow{3}{*}{Ours} & \# LLM call    & \textcolor{green}{1.3}+\textcolor{red}{1.1}  & \textcolor{green}{3.6}+\textcolor{red}{2.2}  & \textcolor{green}{5.3}+\textcolor{red}{2.6} \\
 & Run time (s)   & \textbf{2.50}     & \textbf{4.75}     & \textbf{7.60} \\
 & Iterations & 1.3 &3.6 &5.3\\
\bottomrule
\end{tabular}
\end{table}

\section{Conclusion}
\vspace{-0.2cm}

In this paper, we proposed GraSP, a novel subgraph-level reasoning framework that encodes subgraphs into graph-based soft prompts using GNNs. By reducing reliance on explicit knowledge graph edges, GraSP effectively mitigates the impact of missing relations and improves reasoning robustness under incomplete knowledge graphs. We further introduced a two-stage reasoning paradigm, where a lightweight LLM identifies question-relevant entities guided by soft prompts, followed by a more powerful LLM for evidence-aware answer generation. This modular design not only reduces computational cost, but also enables GraSP to be seamlessly integrated with off-the-shelf LLMs. Experiments on four benchmarks demonstrate state-of-the-art performance on three datasets, while extensive analyses and ablation studies further validate the effectiveness and robustness of our approach.

\section*{Limitations}
While our approach demonstrates promising effectiveness, it still has several limitations. First, our method relies on the structural information of the underlying knowledge graph. When the knowledge graph is severely incomplete, contains sparse information, or exhibits overly simple structures, the extracted subgraphs may provide limited useful evidence, which in turn constrains the effectiveness of our approach. Second, our framework employs soft prompts, which are continuous representations without explicit semantic grounding. As a result, the reasoning process becomes less interpretable, making it difficult to precisely understand how specific structural signals influence the LLM’s decision-making.

\bibliography{custom}

\appendix

\begin{table*}[ht]
\centering
\caption{Overview of datasets.}
\label{tab:datasets}
\resizebox{0.8\textwidth}{!}{
\begin{tabular}{@{}llllll@{}}
\toprule
\textbf{Dataset} & \textbf{Train} & \textbf{Test} & \textbf{Complexity} & \textbf{Reasoning} & \textbf{Annotation} \\
\midrule
IKGQA & -  & 200 & Incomplete KG & Multi-hop  & Ans. only \\
CWQ-IKG         & 27,689 & 1,000 & Incomplete KG         & Multi-hop         & Ans. only\\
WebQSP-IKG      & 3,098  & 1,000 & Incomplete KG        & Multi-hop (most) & Ans. only\\
\hline
CWQ         & 27,689 & 3,531 & Complex Qs         & Multi-hop         & SPARQL, Ans. \\
WebQSP      & 3,098  & 1,639 & Moderate Qs        & Multi-hop (most) & SPARQL, Ans. \\
WebQuestion & 3,778  & 2,032 & Simple–Moderate Qs & Single-hop (most) & Ans. only \\
GrailQA     & 35,138 & 1,000 & Complex (varied)   & Multi-hop         & Logic forms, Ans. \\
\bottomrule
\end{tabular}
}
\end{table*}

\begin{table*}[t]
\centering
\small
\caption{Performance comparison on four KBQA datasets. 
Results marked with `*' are taken directly from the corresponding original papers. 
Bold numbers indicate the \textbf{best performance}, while underlined numbers denote the \underline{second-best} (Hits@1).}
\renewcommand{\arraystretch}{1.1}
\begin{tabular}{lcccc}
\toprule
\multirow{2}{*}{\textbf{Method}} 
& \multicolumn{3}{c}{\textbf{Multi-hop reasoning}} 
& \textbf{Generalization} \\
\cmidrule(lr){2-4} \cmidrule(lr){5-5}
 & CWQ & WebQSP & WebQuestion & GrailQA \\
\midrule
EmbedKGQA~\cite{saxena-etal-2022-kgt5}& 55.7 & 67.0& 60.3& 52.2\\
ToG~\cite{sun2024thinkongraph} & 69.5 & 82.1 & 72.8 & 69.4 \\
KBQA-o1~\cite{luokbqa} & 72.0 & 88.3 & \underline{82.5} & 72.9 \\
KG-Agent~\cite{jiang-etal-2025-kg} & 72.2$^{*}$ & 83.3$^{*}$ & -- & \underline{86.1}$^{*}$ \\
iQUEST~\cite{wang-yu-2025-iquest} & 73.8$^{*}$ & 88.9$^{*}$ & 81.2$^{*}$ & 73.5$^{*}$ \\
LMP~\cite{wan2025digest}  & \underline{82.2}$^{*}$ & \underline{90.0}$^{*}$ & 80.4$^{*}$ & \textbf{89.3}$^{*}$ \\
\midrule
\textbf{GraSP (Ours)} & \textbf{82.6} & \textbf{93.6} & \textbf{84.3} & {84.0} \\
\bottomrule
\end{tabular}
\label{tab:kbqa_results_four_widely_used}
\end{table*}

\section{Implementation Details}
\label{sec:impl}
In our main experiments, we use GPT-5.2 as the answer LLM and GPT-OSS-20B as the selection LLM. Specifically, we apply LoRA to the {query and value projection matrices} of each self-attention layer, with the LoRA rank set to {$r=8$} and the scaling factor set to {$\alpha=16$}. A dropout rate of {0.1} is applied to the LoRA layers. During fine-tuning, only the LoRA parameters are updated while all original model parameters remain frozen.

During subgraph construction, we extract 2-hop subgraphs, and accordingly employ a two-layer attention-based GNN for subgraph encoding. We adopt \texttt{bert-base-uncased} as the text encoder, which has a hidden size of 768, while the hidden dimension of the GNN is set to 128. Since GPT-OSS-20B has a hidden size of 2880, the feed-forward mapping layer that projects GNN representations into soft prompts has a dimension of 128 × 2880. All models are trained on 8 NVIDIA A100 GPUs (80GB each). In addition, we conduct extensive comparisons using LLMs of different scales, including Qwen3-0.6B, Qwen3-1.7B, Qwen3-4B, Qwen3-8B, Qwen-30B-A3B, LLaMA-3.3-7B, and GPT-4o.

For constructing ground-truth labels for question-related entity selection (i.e., $y$ in Function~\ref{func:loss}), we treat the answer entity as the most relevant entity for 1-hop questions. For multi-hop questions, all entities along the shortest path between the topic entity and the answer entity are considered relevant.

\section{Datasets}
\label{app-datasets}
The statistics of these four multi-hop KBQA datasets used in this work are shown in Table \ref{tab:datasets}, including the number of training and test instances, question complexity, required reasoning types, and available annotation formats.

\label{app:dataset}

\section{SPARQL Query}
\label{sec:sparql}
As the SPARQL syntax for retrieving neighboring entities is fixed, a common template can be reused to systematically obtain such information. For example, to retrieve all cities located in Germany, we issue the following SPARQL query:
\begin{lstlisting}[style=sparql]
SELECT ?tailEntity
WHERE {
  ns:m.0345h ns:location.country.cities ?tailEntity .
}
\end{lstlisting}
\noindent
where \texttt{ns:m.0345h} is the Freebase ID for Germany, and \texttt{ns:location.country.cities} denotes the corresponding predicate.

\begin{table}[t]
\centering
\small
\caption{Comparison with different baselines.}
\renewcommand{\arraystretch}{1.1}
\begin{tabular}{lcc}
\toprule
\textbf{Method} 
 & \textbf{CWQ} & \textbf{WebQSP} \\
\midrule
RoG \cite{luoreasoning} & 62.6$^{*}$ & 85.7$^{*}$ \\
GNN-RAG \citeyearpar{mavromatis2024gnn} & 62.8$^{*}$ & 82.8$^{*}$ \\
PoG \cite{tan2025paths} & 65.1 & 86.2 \\
LightPROF \cite{ao2025lightprof} & 59.3$^{*}$ & 83.8$^{*}$ \\
MemQ \cite{xu2025memory} & 80.3$^{*}$ & 84.1$^{*}$ \\
KaeDe \cite{bu2025kaede} & 88.3$^{*}$ & 91.1$^{*}$ \\
ChatKBQA \cite{luo2024chatkbqa}& 86.4$^{*}$ & 86.0$^{*}$ \\
\midrule
\textbf{GraSP (Ours)} & 82.6 & \textbf{93.6} \\
\bottomrule
\end{tabular}
\label{tab:kbqa_results_only_two}
\end{table}

\begin{table*}[t]
\centering
\small
\caption{Performance comparison of different select LLMs.}
\renewcommand{\arraystretch}{1.1}
\begin{tabular}{l|cccc}
\toprule
\textbf{Select LLM} & \textbf{CWQ} & \textbf{WebQSP} & \textbf{WebQuestion} & \textbf{GrailQA} \\
\midrule
Qwen3-0.6B & 61.73 & 69.14 & 58.05 & 62.48 \\
\rowcolor{gray!20} Qwen3-0.6B (Finetune) & 71.31 & 81.14 & 68.52 & 69.87 \\
Qwen3-1.7B & 66.24 & 74.53 & 60.72 & 63.28 \\
\rowcolor{gray!20} Qwen3-1.7B (Finetune) & 73.40 & 84.51 & 71.96 & 72.58 \\
Qwen3-4B & 69.66 & 79.68 & 69.01 & 63.09 \\
\rowcolor{gray!20} Qwen3-4B (Finetune) & 77.07 & 86.83 & 76.52 & 77.16 \\
Qwen3-8B & 72.55 & 81.54 & 72.31 & 65.38 \\
\rowcolor{gray!20} Qwen3-8B (Finetune) & 80.85 & 88.86 & 78.40 & 79.16 \\
Qwen-30B-A3B & 76.39 & 88.71 & 78.27 & 77.53 \\
\rowcolor{gray!20} Qwen-30B-A3B (Finetune) & 82.28 & 93.16 & 84.07 & 83.46 \\
GPT-20B & 77.21 & 86.34 & 78.28 & 76.13 \\
\rowcolor{gray!20} GPT-20B (Finetune) & 82.58 & 93.63 & 84.32 & 83.99 \\
\bottomrule
\end{tabular}
\label{tab:different_select_llm}
\end{table*}

\begin{table*}[t]
\centering
\small
\caption{Performance comparison of different answer LLMs.}
\renewcommand{\arraystretch}{1.1}
\begin{tabular}{l|cccc}
\toprule
\textbf{Answer LLM} & \textbf{CWQ} & \textbf{WebQSP} & \textbf{WebQuestion} & \textbf{GrailQA} \\
\midrule
Qwen3-8B & 54.58 & 68.97 & 64.88 & 52.17 \\
Qwen-30B-A3B & 61.91 & 75.65 & 69.49 & 64.18 \\
GPT-oss-20b & 67.90 & 79.98 & 71.74 & 72.88 \\
LLaMA3.3-70B & 68.81 & 79.76 & 74.97 & 70.82 \\
GPT-oss-120b & 75.03 & 84.75 & 79.42 & 78.01 \\
GPT-4o & 77.64 & 90.95 & 82.39 & 79.82 \\
GPT-5.2 & 82.58 & 93.63 & 84.32 & 83.99 \\
\bottomrule
\end{tabular}
\label{tab:answer_llm_results}
\end{table*}

\section{Additional Experiments }
\label{sec:additional_results}

\subsection{Results on Four Widely Used KBQA Datasets}
\label{sec:additional_results_four_widely_used}
The experimental results on four widely used KBQA datasets are reported in Table~\ref{tab:kbqa_results_four_widely_used}. As shown, our approach achieves the best performance on three multi-hop datasets. On the generalization dataset, our results are also comparable to the strongest baseline. Among the compared methods, ToG exhibits the poorest performance, primarily because it relies solely on node-level semantic information at each reasoning step. In contrast, KBQA-o1, iQUEST, and KG-Agent show consistent improvements, indicating that incorporating sub-questions or memory mechanisms can effectively enhance reasoning performance. Notably, KG-Agent extracts entities from both the training and test sets in advance, significantly reducing the search space for the LLM. This substantially lowers the difficulty of evaluating generalization on GrailQA, leading to relatively strong results. 

We also observe that LMP, which also uses subgraphs to enhance reasoning, achieves the best performance among the baselines, as it can utilize richer contextual information at each reasoning step. However, this approach incurs substantial computational overhead due to repeated subgraph extraction and transformation. Also, the transformation process is less effective because introduce noises, and overly large subgraphs can exceed the LLM’s effective context window, negatively affecting performance. In contrast, our method adopts a soft-prompt-based design and employs a GNN to efficiently compute graph representations in parallel, resulting in a more effective and scalable utilization of graph structure.

\subsection{Comparison with Additional Baselines}

Since most previous methods were evaluated on the standard KBQA benchmarks CWQ and WebQSP, we further compare our approach with recent state-of-the-art methods on these two datasets, as shown in Table~\ref{tab:kbqa_results_only_two}. Most baseline results are directly taken from the original papers. Notably, PoG~\cite{tan2025paths} reports results on a randomly selected subset of 1,000 examples rather than the full benchmarks, making direct comparison inappropriate. To ensure fairness, we re-evaluate PoG on the complete datasets, where its performance remains below ours. The results show that our method achieves the best performance on WebQSP and remains highly competitive with the strongest methods on CWQ.

\section{Complete Experimental Results of Two-Stage Reasoning}
\label{app-two-stage}
\paragraph{Evaluate select LLMs}
Different models are evaluated as the select LLMs, including varying model sizes and whether fine-tuning is applied. The results are shown in Table~\ref{tab:different_select_llm}.

\paragraph{Evaluate answer LLMs} 
The effectiveness of using different models as the select LLMs is evaluated, with the results shown in Table~\ref{tab:answer_llm_results}.

\section{Case Study}

\paragraph{System prompt:}
{\ttfamily 
You are a knowledge-graph reasoning assistant.

The user asks a natural-language quesion. You are given a set of retrieved knowledge-graph triples and a graph topology around the entities mentioned in the question. Some retrieved triples may be directly relevant, indirectly useful, or irrelevant.

Your task is to analyze the retrieved graph evidence and identify the reasoning paths that are useful for answering the question. The direct relation needed to answer the question may be missing from the graph, so you should use multi-hop reasoning over the available triples to infer the answer.

You should not simply return all retrieved triples. Instead, you should filter the graph evidence, select the most relevant paths, and summarize how these paths support the answer. Your output should help another agent generate the final answer to the user’s question.

When reasoning, follow these principles:
    
    1. Identify the target relation implied by the question.

2. Check whether a direct edge exists in the knowledge graph. If the direct edge is missing, search for multi-hop paths that can support the missing relation.

3. Prefer paths that are semantically consistent and repeatedly supported by multiple pieces of evidence.

4. Ignore graph triples that do not help answer the question.

5. Produce a concise evidence summary.
}

\subsection{Case 1: One-Iteration Reasoning}
\paragraph{Question:} \texttt{Which prison island was Nelson Mandela strongly associated with?}

The expected answer: \texttt{Robben Island.}

{\color[gray]{0.5} \textit{The knowledge graph does not contain the direct edge}:
\[
\text{Nelson Mandela} \xrightarrow{\text{imprisonedAt}} \text{Robben Island}.
\]
\vspace{-1em}
}
\paragraph{The retrieved first-hop triples:}

{\ttfamily 
\[
\begin{aligned}
\text{Nelson Mandela} &\xrightarrow{\text{fellowPrisoner}} \text{Walter Sisulu}, \\
\text{Nelson Mandela} &\xrightarrow{\text{fellowPrisoner}} \text{Govan Mbeki}, \\
\text{Nelson Mandela} &\xrightarrow{\text{fellowPrisoner}} \text{Raymond Mhlaba}, \\
\text{Nelson Mandela} &\xrightarrow{\text{defendantIn}} \text{Rivonia Trial}.
\end{aligned}
\]
}

\paragraph{The retrieved second-hop triples:}
{\ttfamily 
\[
\begin{aligned}
\text{Walter Sisulu} &\xrightarrow{\text{imprisonedAt}} \text{Robben Island}, \\
\text{Govan Mbeki} &\xrightarrow{\text{imprisonedAt}} \text{Robben Island}, \\
\text{Raymond Mhlaba} &\xrightarrow{\text{imprisonedAt}} \text{Robben Island}, \\
\text{Rivonia Trial} &\xrightarrow{\text{resultedIn}} \text{imprisonment}.
\end{aligned}
\]
}

\paragraph{GNN-encoded two-hop graph structure:}
{\ttfamily 
\[
\begin{aligned}
\text{Nelson Mandela} &\xrightarrow{\text{fellowPrisoner}} \text{Walter Sisulu}, \\
\text{Walter Sisulu} &\xrightarrow{\text{imprisonedAt}} \text{Robben Island}; \\
\text{Nelson Mandela} &\xrightarrow{\text{fellowPrisoner}} \text{Govan Mbeki}, \\
\text{Govan Mbeki} &\xrightarrow{\text{imprisonedAt}} \text{Robben Island}; \\
\text{Nelson Mandela} &\xrightarrow{\text{fellowPrisoner}} \text{Raymond Mhlaba}, \\
\text{Raymond Mhlaba} &\xrightarrow{\text{imprisonedAt}} \text{Robben Island}; \\
\text{Nelson Mandela} &\xrightarrow{\text{defendantIn}} \text{Rivonia Trial}, \\
\text{Rivonia Trial} &\xrightarrow{\text{resultedIn}} \text{imprisonment}.
\end{aligned}
\]

The target answer type is a prison island. Then check whether the direct edge exists in the retrieved graph:
\[
\text{Nelson Mandela} \xrightarrow{\text{imprisonedAt}} \text{? prison island}.
\]

I cannot answer the question using a one-hop lookup. Instead, and need to infer the answer from multi-hop graph evidence. Among the retrieved triples, the most useful reasoning paths are:

Nelson Mandela --> fellowPrisoner --> Walter Sisulu 
--> imprisonedA --> Robben Island; 

Nelson Mandela --> fellowPrisoner --> Govan Mbeki
 --> imprisonedAt --> Robben Island; 
 
Nelson Mandela --> fellowPrisoner --> Raymond Mhlaba
--> imprisonedAt --> Robben Island.

These paths are highly relevant because they connect Nelson Mandela to multiple fellow prisoners who were all imprisoned at the same location, Robben Island. The repeated occurrence of Robben Island across multiple independent two-hop paths provides strong graph evidence that Nelson Mandela was also strongly associated with Robben Island.

The Rivonia Trial path provides additional contextual support:

Nelson Mandela --> defendantIn --> Rivonia Trial
--> resultedIn --> imprisonment.
}

\vspace{3mm}
\noindent
\textbf{Small-LLM agent output:}

{\ttfamily 
This path explains why imprisonment is relevant to Nelson Mandela in this graph neighborhood, although it does not directly identify the prison island. Therefore, it is treated as supporting context rather than the main answer path.

Based on the question, the retrieved triples, and the graph topology, the useful reasoning paths are:

Nelson Mandela --> fellowPrisoner --> Walter Sisulu
 --> imprisonedAt --> Robben Island; 
Nelson Mandela --> fellowPrisoner --> Govan Mbeki
 --> imprisonedAt --> Robben Island;
 Nelson Mandela --> fellowPrisoner --> Raymond Mhlaba
 --> imprisonedAt --> Robben Island.

Although the direct edge from Nelson Mandela to Robben Island is missing, multiple two-hop paths connect Nelson Mandela to Robben Island through his fellow prisoners. Therefore, the most likely answer candidate is Robben Island.
}

\vspace{3mm}
\noindent
\textbf{Powerful-LLM agent output:}

{\ttfamily 
The knowledge graph does not provide a direct edge stating that Nelson Mandela was imprisoned at Robben Island. However, the graph contains several strong multi-hop reasoning paths. Nelson Mandela is connected to Walter Sisulu, Govan Mbeki, and Raymond Mhlaba through the relation fellowPrisoner. Each of these fellow prisoners is connected to Robben Island through the relation imprisonedAt. These repeated paths provide strong evidence that Nelson Mandela was strongly associated with Robben Island. The Rivonia Trial path further supports the imprisonment context, although it does not directly identify the prison island. Therefore, the answer is \{Robben Island\}.
}

\subsection{Case 2: Two-Iteration Reasoning}

\paragraph{Question:}
\texttt{What language do the people in the area where the newspaper Knews is circulated speak?}

The expected answer: \texttt{Greek and Turkish.} 

{\color[gray]{0.5} \textit{The knowledge graph does not contain the direct edge}:
\[
\text{Knews} \xrightarrow{\text{published\_in}} \text{Cyprus}.
\]
\vspace{-1em}
}

\vspace{3mm}
\noindent
{\color[gray]{0.5} \textit{First Iteration:}}

\paragraph{The retrieved first-hop triples:}
{\ttfamily 
\[
\begin{aligned}
\text{Knews} &\xrightarrow{\text{is\_edition\_of}} \text{Kathimerini}, \\
\text{Knews} &\xrightarrow{\text{editor\_in\_chief}} \text{S. B. Kyriakides}, \\
\text{Knews} &\xrightarrow{\text{owned\_by}} \text{SPP\_Media\_Group}
\end{aligned}
\]
}

\paragraph{The retrieved second-hop triples:}
{\ttfamily 
\[
\begin{aligned}
\text{S. B. Kyriakides} &\xrightarrow{\text{nationality}} \text{Cyprus}, \\
\text{S. B. Kyriakides} &\xrightarrow{\text{employer}} \text{SPP\_Media\_Group}, \\
\text{S. B. Kyriakides} &\xrightarrow{\text{born\_in}} \text{Phoenix}, \\
\text{SPP\_Media\_Group} &\xrightarrow{\text{operates\_in}} \text{Cyprus}, \\
\text{SPP\_Media\_Group} &\xrightarrow{\text{CEO}} \text{Dimitris\_Lottides}, \\
\text{SPP\_Media\_Group} &\xrightarrow{\text{founded\_year}} 1999, \\
\text{Kathimerini} &\xrightarrow{\text{has\_regional\_focus}} \text{Cyprus}, \\
\text{Kathimerini} &\xrightarrow{\text{has\_regional\_focus}} \text{Greece}, \\
\text{Kathimerini} &\xrightarrow{\text{founder}} \text{Georgios\_Vlahos}.
\end{aligned}
\]
}

\paragraph{GNN-encoded two-hop graph structure:}
{\ttfamily 
\[
\begin{aligned}
\text{Knews} &\xrightarrow{\text{is\_edition\_of}} \text{Kathimerini}, \\
\text{Knews} &\xrightarrow{\text{editor\_in\_chief}} \text{S. B. Kyriakides}, \\
\text{Knews} &\xrightarrow{\text{owned\_by}} \text{SPP\_Media\_Group}, \\ 
\text{S. B. Kyriakides} &\xrightarrow{\text{nationality}} \text{Cyprus}, \\
\text{S. B. Kyriakides} &\xrightarrow{\text{employer}} \text{SPP\_Media\_Group}, \\
\text{SPP\_Media\_Group} &\xrightarrow{\text{operates\_in}} \text{Cyprus}, \\
\text{Kathimerini} &\xrightarrow{\text{has\_regional\_focus}} \text{Cyprus}, \\
\end{aligned}
\]
}

\vspace{3mm}
\noindent
\textbf{Small-LLM agent output:}
{\ttfamily 
The main entity in the question is Knews, and the target answer type is a language. The phrase ``the area where the newspaper Knews is circulated'' implies that the agent should first identify the geographic area associated with Knews. In the knowledge graph, the closest operational relation corresponding to this intent is:
"Knews -->  published\_in / circulated\_in / operates\_in --> ? area."
The agent then checks whether the direct edge exists in the retrieved graph. Since the direct triple
"Knews --> published\_in -->  Cyprus" is missing from the incomplete graph, the agent cannot immediately identify the circulation area using a one-hop relation. Instead, it needs to infer the area from multi-hop graph evidence.
Among the retrieved first-round triples, the most useful reasoning paths are:

Knews -->  owned\_by -->  SPP\_Media\_Group,
 --> operates\_in -->  Cyprus, \\
Knews -->  is\_international\_edition\_of --> Kathimerini
--> has\_regional\_focus -->  Cyprus.

These paths are highly relevant because they connect Knews to Cyprus through organizational and regional-focus evidence. The path through SPP Media Group indicates that the owner of Knews operates in Cyprus. The path through Kathimerini indicates that the newspaper's related publication has a regional focus on Cyprus. Together, these paths provide strong graph evidence that the area associated with Knews is Cyprus.
}

\vspace{3mm}
\noindent
\textbf{Powerful-LLM agent output:}

{\ttfamily 
Although the direct edge from Knews to Cyprus is missing, multiple two-hop paths connect Knews to Cyprus through relevant organizational and regional relations. Therefore, the next reasoning step should be grounded on the entity Cyprus.

In the second round of retrieval, the agent queries the graph neighborhood around \{Cyprus\}. 
}

\vspace{3mm}
\noindent
{\color[gray]{0.5} \textit{Second Iteration:}}

\paragraph{The retrieved first-hop triples:}
\[
\begin{aligned}
\text{Cyprus} &\xrightarrow{\text{official\_language}} \text{Greek}, \\
\text{Cyprus} &\xrightarrow{\text{official\_language}} \text{Turkish}, \\
\text{Cyprus} &\xrightarrow{\text{capital}} \text{Nicosia}.
\end{aligned}
\]

\paragraph{The retrieved second-hop triples:}
\[
\begin{aligned}
\text{Greek} &\xrightarrow{\text{language\_family}} \text{Indo-European languages}, \\
\text{Greek} &\xrightarrow{\text{writing\_system}} \text{Greek alphabet}, \\
\text{Greek} &\xrightarrow{\text{native\_to}} \text{Greece}, \\
\text{Turkish} &\xrightarrow{\text{language\_family}} \text{Turkic languages}, \\
\text{Turkish} &\xrightarrow{\text{writing\_system}} \text{Latin alphabet}, \\
\text{Turkish} &\xrightarrow{\text{official\_language\_of}} \text{Turkey}, \\
\text{Nicosia} &\xrightarrow{\text{country}} \text{Cyprus}, \\
\text{Nicosia} &\xrightarrow{\text{is\_capital\_of}} \text{Cyprus}, \\
\text{Nicosia} &\xrightarrow{\text{located\_in}} \text{Nicosia District}.
\end{aligned}
\]

\paragraph{GNN-encoded two-hop graph structure:}
{\ttfamily 
\[
\begin{aligned}
\text{Cyprus} &\xrightarrow{\text{official\_language}} \text{Greek}, \\
\text{Cyprus} &\xrightarrow{\text{official\_language}} \text{Turkish}, \\
\text{Greek} &\xrightarrow{\text{language\_family}} \text{Indo-European}, \\
\text{Greek} &\xrightarrow{\text{writing\_system}} \text{Greek alphabet}, \\
\text{Greek} &\xrightarrow{\text{native\_to}} \text{Greece}, \\
\text{Turkish} &\xrightarrow{\text{language\_family}} \text{Turkic languages}, \\
\text{Turkish} &\xrightarrow{\text{writing\_system}} \text{Latin alphabet}, \\
\text{Turkish} &\xrightarrow{\text{official\_language\_of}} \text{Turkey}.
\end{aligned}
\]
}

\vspace{3mm}
\noindent
\textbf{Small-LLM agent output:}

{\ttfamily 
The question asks what language people in the area speak. Therefore, the relevant relation is:
Cyprus --> official\_language --> ? language.

Based on the question, the retrieved triples, and the graph topology, the useful reasoning paths are:
 Cyprus -->  official\_language -->  Greek, \\
 Cyprus -->  official\_language -->  Turkish 

The official languages of Cyprus are Greek and Turkish. Therefore, the answer candidate is Greek and Turkish.
}

\vspace{3mm}
\noindent
\textbf{Powerful-LLM agent output:}

{\ttfamily 
 The knowledge graph does not provide a direct edge stating that Knews is published in or circulated in Cyprus. However, the graph contains several strong multi-hop reasoning paths connecting Knews to Cyprus. Knews is owned by SPP Media Group, and SPP Media Group operates in Cyprus. Knews is also the international edition of Kathimerini, which has a regional focus on Cyprus. These paths provide strong evidence that the relevant circulation area for Knews is Cyprus. Once Cyprus is identified as the area, the second-round retrieval shows that Cyprus has two official languages: Greek and Turkish. Therefore, the people in the area where Knews is circulated speak \{Greek\} and \{Turkish\}.
}

\end{document}